\documentclass[12pt]{article}

\usepackage{sbc-template}
\usepackage{graphicx,url}
\usepackage[utf8]{inputenc}
\usepackage[brazil]{babel}

\title{True to Tone? Quantifying Skin Tone Fidelity and Bias in Photographic-to-Virtual Human Pipelines}

\author{Gabriel Ferri Schneider\inst{1}\inst{4}, Erick Menezes\inst{2}\inst{4}, Rafael Mecenas\inst{2}\inst{4}, Paulo Knob\inst{1},  \\ Victor Araujo\inst{1}\inst{2}\inst{3}\inst{4}, Soraia Musse\inst{1}\inst{4}}

\address{PUCRS\\
  Porto Alegre -- RS -- Brazil
\nextinstitute
  UNIT\
  Araujo -- SE -- Brazil
\nextinstitute
  Kunumi Institute\\
  Brazil
  \nextinstitute
  INCT-SANI\\
  Brazil
}

\begin{document} 

\maketitle

\begin{abstract}
  Accurate reproduction of facial skin tone is essential for realism, identity preservation, and fairness in Virtual Human (VH) rendering. However, most accessible avatar creation pipelines rely on photographic inputs that lack colorimetric calibration, which can introduce inconsistencies and bias. We propose a fully automatic and scalable methodology to systematically evaluate skin tone fidelity across the VH generation pipeline. Our approach defines a full workflow that integrates skin color and illumination extraction, texture recolorization, real-time rendering, and quantitative color analysis. Using facial images from the Chicago Face Database (CFD), we compare skin tone extraction strategies based on cheek-region sampling, following the literature, and multidimensional masking derived from full-face analysis. Additionally, we test both strategies with lighting isolation, using the pre-trained TRUST framework, employed without any training or optimization within our pipeline. Extracted skin tones are applied to MetaHuman textures and rendered under multiple lighting configurations. Skin tone consistency is evaluated objectively in the CIELAB color space using the $\Delta E$ metric and the Individual Typology Angle (ITA). The proposed methodology operates without manual intervention and, with the exception of pre-trained illumination compensation modules, the pipeline does not include learning or training stages, enabling low computational cost and large-scale evaluation. Using this framework, we generate and analyze approximately 19,848 rendered instances. Our results show phenotype-dependent behavior of extraction strategies and consistently higher colorimetric errors for darker skin tones.
\end{abstract}
     
\begin{figure}
  \centering
  \includegraphics[width=16cm]{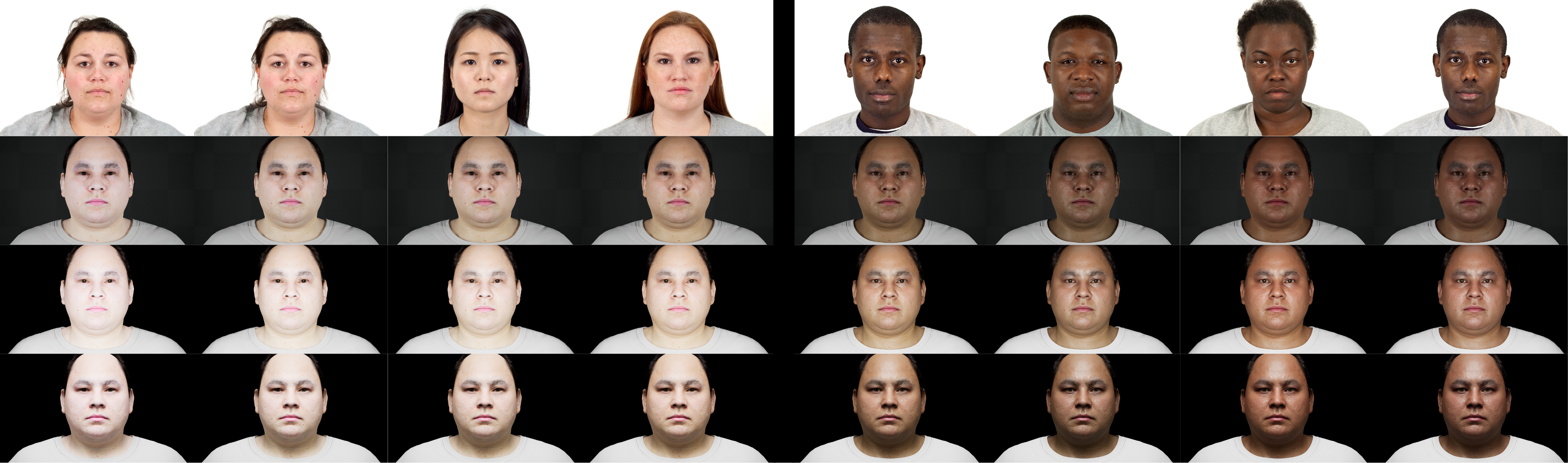}
  \caption{From left to right, the first four columns represent subjects from the CFD dataset classified as ITA Class 1 (lighter skin tones), while the second group of four columns represents ITA Class 6 (darker skin tones). Each column corresponds to one extraction method for each group: $Cheek$, $MMM$, $T-Cheek$, and $T-MMM$. Additionally, each row (from top to bottom) shows the results for a different lighting configuration: CFD, Frontal, and Paramount.}
  \label{fig:teaser}
\end{figure}

\section{Introduction}

Skin tone is a complex and multifaceted characteristic, the perception of which varies across cultural, social, and contextual boundaries~\cite{canache2014determinants,kang2015multiple,fuhrer2021social,brunsma2001new}. What is considered ``light'' or ``dark'' skin is not solely determined by physical pigmentation, but is also shaped by social constructs and lived experiences. This inherent subjectivity poses significant challenges for the faithful representation of skin tones in digital environments. As Virtual Humans (VHs) become increasingly prevalent in games, films, and immersive applications, the demand for personalized avatars generated from personal photographs has grown substantially. In this context, ensuring that skin tones are processed and rendered with equal fidelity is both a technical challenge and an ethical imperative in modern Computer Graphics (CG).

Historically, CG research and production tools have disproportionately used light skin tones as a default reference, resulting in biased rendering workflows. Kim et al.~\cite{kim2022countering} identified an overdependence on subsurface scattering parameters tuned for lighter skin, which often leads to the over-lightening and misrepresentation of darker skin tones. In parallel, extracting accurate skin color from photographic inputs remains challenging, particularly when images lack colorimetric calibration. Feng et al.~\cite{feng2022fairness} demonstrated that scene ambiguities and illumination effects significantly impact 
unbiased albedo estimation.
These technical limitations have perceptual consequences: Araujo et al.~\cite{de2023evaluating} showed that VHs with darker skin tones are consistently perceived as less realistic across media, reinforcing disparities associated with the Uncanny Valley effect. Altogether, these findings underscore the need to examine how modern pipelines propagate errors and biases from photographic inputs to rendered VHs.

Some studies aimed to improve how skin tone is categorized and analyzed. While the Fitzpatrick scale remains widely used, it has been criticized for its limited range of representation, particularly for darker skin tones. More inclusive alternatives~\cite{heldreth2024skin}, such as the Monk Skin Tone scale~\cite{monk2019monk} and multidimensional color representations that incorporate hue and undertone information~\cite{thong2023beyond}, have highlighted the importance of moving beyond simple lightness-based descriptors. However, improved descriptors alone are insufficient if rendering pipelines lack systematic ways to evaluate how skin tone information is extracted, transformed, and rendered across diverse conditions.
To address this gap, we adopted the Individual Typology Angle (ITA)~\cite{chardon1991skin}, computed directly from CIELAB values, to objectively categorize skin phenotypes into six groups ranging from Very Light to Dark, without relying on subjective annotations. Using facial photographs from the Chicago Face Database (CFD)~\cite{ma2015chicago,ma2021chicago,lakshmi2021india}, 
which provides controlled yet non-color-calibrated imagery representative of faces,
we investigate the following central question: \emph{what is the most appropriate methodology for reproducing the skin tone of a specific photograph in a realistic VH?} Rather than evaluating isolated techniques, we examine how choices made at different stages of the pipeline interact to affect color fidelity and bias.

In this work, we present a fully automatic, scalable methodology that enables systematic evaluation of skin tone consistency throughout the VH generation pipeline. Our approach integrates skin color and illumination extraction, texture re-colorization, real-time rendering, and quantitative color analysis within a single framework. 
Our goal is to investigate the following research questions: \textbf{(Q1) Extraction Impact:} How do different skin color extraction strategies, based on cheek regions or full-face multidimensional masking, and the inclusion of illumination compensation affect the final rendering, measured by $\Delta E$ and ITA?; \textbf{(Q2) Phenotype Sensitivity:} How does skin tone reproduction error vary across different ITA-based skin phenotype
classifications?; \textbf{(Q3) Lighting Influence:} To what extent do different rendering environments, including frontal, Paramount, and image-based lighting derived from the input images, influence error propagation? 
To answer these questions, we render MetaHuman characters in Unreal Engine\footnote{https://www.unrealengine.com} under multiple lighting configurations and evaluate the results objectively using $\Delta E$ and ITA-based error measures in the CIELAB color space. The fully automatic nature of the pipeline enables large-scale experimentation: using this methodology, we generate and analyze approximately $19,848$ rendered instances. This scale allows us to assess not only average color fidelity, but also how errors vary across extraction methods, skin phenotypes defined by ITA, and lighting conditions.

\section{Related Work}
\label{sec:relatedWork}

Recently, there has been much discussion about the skin color used in the embodiment of virtual agents, a discussion echoed by psychophysical research. Bartlett and Krogmeier~\cite{bartlett2025observations} point out that most Virtual Reality studies match avatars to participants’ gender, but not their skin color, effectively excluding considerations of skin tone in research on embodiment and bias. VHs are present in various media, while developers are able to generate more and more perfect skin and hair; however, mostly for white skin tones~\cite{kim2020racist,kim2022countering}. In fact,~\cite{kim2022countering} discussed the racial bias that influences technical aspects of research, proposing new practices for a more comprehensive approach to CG research. ~\cite{feng2022fairness,feng2022towards} comment that the majority of methods used to estimate facial appearance are biased for light skin tones. \cite{malazita2022using} questions the assumptions used in rendering workflows, especially for dark skin tones. Malazita goes further, challenging the CG community to re-examine core rendering assumptions—especially skin tone—so that virtual representations better reflect real-world diversity and avoid reinforcing inequities. Schumann et al.~\cite{schumann2023consensus} proposed a skin tone scale designed for skin tone classification, which is significantly more inclusive than the
previous classification method, such as the Fitzpatrick Skin Type~\cite{fitzpatrick1975soleil}. 

In the entertainment industry, artists have recognized the need for customized solutions. The team at Walt Disney Animation, for example, updated its skin shaders and lighting workflows to faithfully render a wide spectrum of skin types in Strange World~\cite{khoo2023lighting}. Earlier studies found that smooth, “pure” skin renderings were judged younger and more appealing than those with pronounced pigmentation variation~\cite{zell2019perception}. Meanwhile, algorithmic bias has been observed in emerging AI-based renderers.~\cite{zeng2025analyzing} found that a state-of-the-art relightable face generator struggled to maintain darker skin tones under varying lighting conditions and tended to produce lighter skin albedos without intervention.~\cite{wisessing2024blinded} showed that using only portrait lighting design can help to change the perception of a VH’s realism and believability in a significant way. Zell et al.~\cite{zell2015stylize} reported that physically accurate shading improves realism across stylization levels, whereas dark cast shadows systematically lower appeal. 

Johnson et al.~\cite{johnson2023deep} modeled age and emotion-dependent skin color by training a spatially-aware autoencoder that edits melanin and haemoglobin parameter maps in latent space; landmark-guided texture transfer and importance-sampled chromophore priors let the method generalize across the full Fitzpatrick tone range. \cite{liu2025controllable} presented a diffusion model for synthesizing biophysically plausible faces and editing texture maps of captured 3D faces based on high and low-level parameters, such as age and melanin level.
~\cite{robin2020beyond} showed that neural networks could achieve superior performance in skin tone estimation, while~\cite{aliaga2023hyperspectral} proposed an encoder-decoder network for estimating the spectral parameters of a face in the visible and near infrared spectra.
%
Taken together, these results underline that both skin accuracy and illumination strategies are pivotal for creating diverse, believable VHs. These findings motivate our investigation into how a modern real-time rendering engine handles skin tone.

\section{Methodology}
\label{sec:method}

\begin{figure*}[!htb]
  \centering
  \includegraphics[width=0.9\textwidth]{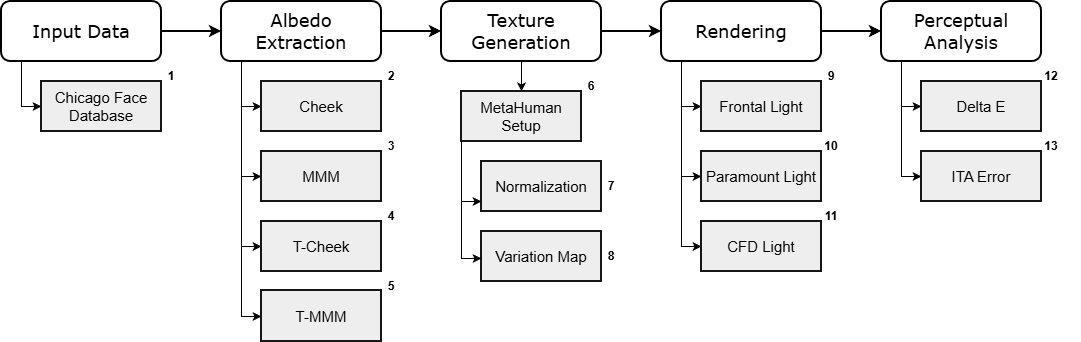}
  \caption{Overview of our methodology for evaluating skin tone consistency. Starting with the ground truth images extracted from the CFD Dataset~\cite{ma2015chicago,ma2021chicago,lakshmi2021india} (1), we extract the average skin color following four different approaches: $Cheek$ (2), $MMM$ (3), $T-Cheek$ (4), and $T-MMM$ (5). Considering our MetaHuman Setup (6), we apply a normalization (7) and a variation map (8) in the texture, before rendering the final image with three different lighting configurations: Frontal Light (9), Paramount Light (10), and CFD Light (11).
  Finally, we validate the resulting rendered images with two metrics: $\Delta E$ (12) and ITA Error (13).}
  \label{fig:overview-model}
\end{figure*}

The overall methodology is illustrated in Figure~\ref{fig:overview-model}. Our goal is to assess the fidelity and fairness of the rendering pipeline when translating a 2D photograph into a 3D VH. 
All stages of the proposed workflow are fully automated and parameterized, enabling exact reproduction of experiments with the same input images and rendering settings, and are detailed in the next sections. 

\subsection{Input Data - Chicago Face Database (CFD)}
\label{sec:CFD}

As presented in Figure~\ref{fig:overview-model} (1), the input dataset is the Chicago Face Database (CFD)~\cite{ma2015chicago}, which provides high-resolution standardized photographs of male and female individuals aged between 17 and 65 years, taken under controlled lighting conditions and with neutral facial expressions. This dataset is particularly suitable for this study due to its broad demographic coverage. The main dataset comprises 597 images of unique individuals who self-identified as Asian, Black, Latino, and White. In addition, we incorporated the CFD-MR extension dataset~\cite{ma2021chicago}, which includes 88 individuals of multiracial ancestry, alongside the India Face Set~\cite{lakshmi2021india}, which includes 142 images of people recruited in Delhi, India, totaling 827 images. 

\subsection{Albedo Extraction Methods}
\label{sec:colorExtraction}

\subsubsection{Cheek-based Extraction - Figure~\ref{fig:overview-model} (2).}
\label{sec:methodRealImage}

Our goal in this strategy (referred to as $Cheek$) is to compute the average sRGB values from input facial images, focusing on the cheek regions, which were chosen for their relatively uniform skin texture and color-representative characteristics~\cite{yan2020exploring,menezes2025evaluating}. All color values are consistently handled in the sRGB color space, assuming standard sRGB primaries, D65 white point, and the sRGB transfer function. No attempt is made to recover absolute spectral reflectance, as the goal of this work is to evaluate practical workflows operating on photographic sRGB inputs. 
To accomplish this, we used OpenCV\footnote{https://opencv.org/} and Haar-like features~\cite{viola2001rapid} for face detection. Once a face was detected, we identified two cheek regions located in the lower part of the face, one on the left and one on the right. Each cheek region was defined as a square with a width equal to 15\% of the horizontal size of the detected face. These squares were then positioned using a horizontal offset of 18\% of the face width from the left and right boundaries, respectively. For vertical placement, we applied an offset equal to 50\% of the vertical face height from the top of the face, and positioned the square regions accordingly. 
Although this approach reflects common practices in digital content creation workflows, it remains sensitive to lighting conditions and the inherent light-albedo ambiguity of unrestricted photography. For this reason, we subsequently evaluated variants that explicitly decouple lighting from intrinsic skin properties using intrinsic decomposition.

\subsubsection{Multidimensional Masking Method - Figure~\ref{fig:overview-model} (3).}
\label{sec:sony}

Following the multidimensional masking strategy proposed by Thong et al.~\cite{thong2023beyond}, we adopted a statistically driven color sampling approach, designed to reduce the influence of local lighting artifacts on skin color estimation from unconstrained photographs. To simplify the nomenclature, we refer to this method as the Multidimensional Masking Method ($MMM$). 
Unlike region-of-interest methods (e.g., the $Cheek$ method), which are based solely on geometry, this strategy operates directly in color space, filtering pixels based on their multidimensional distributions rather than just their spatial locations.
The $MMM$~\cite{thong2023beyond} operates by segmenting the facial skin region using a dense facial mesh obtained via MediaPipe\footnote{https://chuoling.github.io/mediapipe/}, while defining the facial region of interest by computing the convex hull of the detected landmarks. From the remaining skin pixels, color values are analyzed in the CIELAB space, where the K-means ($k=5$) clustering algorithm is applied to group pixels. To determine the final color, the top 3 clusters with the highest luminosity are selected, and their weighted average is computed.
This method operates in the CIELAB color space converted from sRGB inputs and does not attempt to recover intrinsic reflectance or the material's physical parameters. Thus, although this method substantially improves robustness over naive spatial averaging, it remains sensitive to the global light-albedo ambiguity inherent in photographs captured under unknown lighting conditions.

\subsubsection{TRUST Method - Figure~\ref{fig:overview-model} (4) (5).}
\label{sec:trust}

To mitigate the light influence that could affect color estimation from facial photographs of the CFD dataset, we adopted the TRUST intrinsic decomposition method (referred to as $T$) proposed by Feng et al.~\cite{feng2022fairness,feng2022towards}, which is part of approaches 4 and 5 of our methodological pipeline shown in Figure \ref{fig:overview-model}. 
Standard color sampling techniques implicitly assume that observed pixel values directly correspond to surface reflectance, an assumption that is particularly problematic for skin due to subsurface scattering, self-shading, and non-uniform lighting. This often results in darker skin tones being systematically misinterpreted as regions of lower illumination applied to lighter skin, leading to biased reflectance estimation~\cite{feng2022fairness,feng2022towards}. The $T$ method aims to solve this problem by performing scene-aware inverse rendering that models ambient lighting using spherical harmonics estimated from the entire image context, rather than relying solely on local facial cues. 
By incorporating global scene information, $T$ separates incident illumination from intrinsic surface properties, generating an albedo map that is invariant to lighting conditions and more faithful to the subject's original skin color.
In our pipeline, $T$ is applied to all input images, producing a per-pixel intrinsic albedo map expressed in sRGB space. It is important to note that this albedo representation is treated as a material property rather than an observed color, making it suitable for later use in rendering pipelines, such as Unreal Engine. This step serves as the basis for all subsequent color-extraction strategies evaluated in this work.

\paragraph{TRUST-Cheek Method - Figure~\ref{fig:overview-model} (4).}
\label{sec:trustCheek}

In this approach, we use a $T$ strategy to isolate the effect of illumination (Figure~\ref{fig:overview-model} (4)), preserving the same geometric sampling procedure used in the $Cheek$ baseline (Section~\ref{sec:methodRealImage}). Specifically, the TRUST-Cheek method (here referred to as $T-Cheek$) uses the same cheek-based region-of-interest definition, but operates on an intrinsic albedo map rather than observed sRGB values in image space. In this workflow, each input image is first processed using the $T$ method to separate ambient illumination from intrinsic surface properties and generate a dense albedo map. Sampling of the cheek region is then applied directly to this albedo representation, using the same facial location and the same geometric parameters as the $Cheek$ method. This ensures that any observed differences between the two methods ($Cheek$ and $T-Cheek$) can be attributed solely to the removal of lighting effects, and not to changes in spatial sampling or region selection.

\paragraph{TRUST-MMM Method - Figure~\ref{fig:overview-model} (5).}
\label{sec:trustSony}

The TRUST-MMM (referred to as $T-MMM$) also isolates the effect of intrinsic illumination, preserving the same multidimensional masking procedure defined in the $MMM$ baseline (Section~\ref{sec:sony}). The $MMM$ is applied unchanged, but operates on an albedo representation invariant to illumination, rather than observed sRGB values in image space. Similarly to what was done for $T-Cheek$,
each input image is initially processed using $T$ to separate ambient illumination. The multidimensional masking and clustering procedure is then applied directly to this albedo map, using the same segmentation and clustering parameters and the same selection criteria as the $MMM$. This design ensures that the differences between $MMM$ and $T-MMM$ can be attributed to the removal of illumination effects, and not to changes in sampling or filtering strategy.

\subsection{Texture Generation and Final Image Rendering}

\subsubsection{MetaHuman Setup and Assumptions - Figure~\ref{fig:overview-model} (6).}
\label{sec:methodMH1}

First of all, we considered Unreal Engine to be a production-grade, "black box" rendering system, using its publicly exposed parameters without modification. Our analysis focuses on how different color extraction strategies propagate through this system, rather than on the internal physical corrections of the shader model. We input the sRGB computed in the last phase of the pipeline, the engine internally converts them to Linear RGB for rendering the MetaHumans (MH), and finally remaps the result back to sRGB. This process happens in phase (6) of Figure~\ref{fig:overview-model}. To apply this color in MetaHuman Creator\footnote{https://dev.epicgames.com/documentation/en-us/metahuman/metahuman-creator}, we used a single MH model whose base skin texture is initialized from a neutral reference location in UV space, corresponding to the coordinates $(U, V) = (0.5, 0.5)$ in the texture map.
This choice provides a centered and unbiased starting point for texture recoloring, independent of chromatic information. 

\subsubsection{Texture Recoloring Strategies}
\paragraph{Normalization-Based Recoloring - Figure~\ref{fig:overview-model} (7).}

In the normalization-based strategy, we follow a commonly adopted texture recoloring approach, inspired by intrinsic image manipulation techniques~\cite{li2014intrinsic}. Given a base skin texture $Text_{base}$ generated by MetaHuman Creator, we first calculate its average sRGB value $\mu(Text_{base})$. The texture is then normalized by dividing each texel by this average and subsequently scaled by the target skin color vector $C_{target}$ extracted using one of the methods described in Section~\ref{sec:colorExtraction}. 
This operation preserves the relative spatial variation of the original texture while imposing the desired overall skin tone. We assume that the chromatic variation in the texture can be modeled multiplicatively, and that the base texture is approximately color-neutral. As a result, fine skin details, such as pores and freckles, are maintained, while the overall appearance is adjusted towards the target tone.

\paragraph{Variation Map Recoloring - Figure~\ref{fig:overview-model} (8).}

As an alternative to normalization, we used a variation map strategy that separates global color from local texture variation. In this formulation, we calculated a variation map $V$ by subtracting the average sRGB value of the base texture from each texel $Text_{base}$. The recolored texture is then obtained by adding the target skin color vector $C_{target}$ to this variation map. Unlike normalization-based recoloring, this approach treats texture variation as an additive component, rather than a multiplicative one. This distinction is particularly relevant for darker skin tones, where color values lie predominantly in the low-luminance range of 
non-linear color spaces, such as sRGB. In this situation, multiplicative scaling tends to compress intensity variations, reducing local contrast and suppressing fine-grained skin details, which can negatively impact perceptual fidelity.

\subsubsection{Rendering Conditions - Figure~\ref{fig:overview-model} (9) (10) (11).}

Each recolored texture is applied to the MH model in Unreal Engine 5.6.1, in a project initialized with the Film / Video \& Live Events preset with no settings changed, using an automated scripting pipeline. For each texture, we rendered the character under three distinct lighting configurations. Following the evaluation protocol proposed by Wisessing et al.~\cite{wisessing2024blinded}, we adopted two studio lighting configurations widely used by contemporary content creators: a Frontal and a Paramount lights. For these configurations, the default Cine Camera Actor was placed at 150 cm from the MH face. In addition to these controlled studio conditions, we included a third lighting configuration, named CFD Light (Figure~\ref{fig:overview-model} (11)), that aims to reproduce the ambient lighting of the original photograph. Specifically, we reconstructed the scene's lighting using spherical harmonic coefficients estimated by the $T$ method~\cite{feng2022fairness,feng2022towards}. The same camera settings were used as in~\cite{ma2015chicago}. To compensate for the different zoom level, the camera was moved to 120 cm from the MH.
The images were rendered using the Deferred Renderer on a desktop with an AMD Ryzen 7 5700X, 32GB of RAM, and a Nvidia RTX 5060. 
In the automatic process, we rendered 19,848 images, with an average computational time of 3.7 seconds per image.
In Figure~\ref{fig:teaser}, the second, third, and fourth lines represent CDF Light, Front Light, and Paramount Light, respectively.

\subsection{Objective Perceptual Analysis}
\label{sec:perceptualAna}

To quantitatively assess skin tone fidelity throughout the VH generation process, we analyzed a total of 19,848 rendered images, obtained by combining four skin color extraction strategies, two texture recoloring methods, three lighting configurations, and 827 real facial images from CFD, including their multiracial and Indian subsets.

\subsubsection{Perceptual Color Difference ($\Delta E$) - Figure~\ref{fig:overview-model} (12).}

To measure the perceptual differences between the rendered MetaHumans and their respective reference images, we adopted the CIELAB color space and calculated the color distance metric $\Delta E$ (CIE76). This metric has been widely used in psychophysics and color science studies due to its approximate perceptual uniformity \cite{minaker2021optimizing,johnson2024biophysically}. 
First of all, we use the albedo extraction methods to take sRGB values from the facial images in the CFD and the rendered MetaHuman images. These extracted colors are then converted from sRGB to CIELAB using a D65 white point. This transformation enables a more perceptually uniform comparison, as Euclidean distances in CIELAB better reflect how humans perceive color differences. Next, we calculated the $\Delta E$ between the matching samples to quantify the impact of extraction, recoloring, and rendering conditions on the accuracy of skin tone perception. Formally, $\Delta E$ is calculated as: $\Delta E = \sqrt{(L_2 - L_1)^2 + (a_2 - a_1)^2 + (b_2 - b_1)^2},$ where $L$ represents the lightness component of a color, $a$ and $b$ represent the color components which correspond to the human perception as red/green and yellow/blue, respectively. According to established perceptual thresholds~\cite{minaker2021optimizing,johnson2024biophysically}, the interpretation of $\Delta E$ values is as follows: \textit{i)} $\Delta E \leq 1.0$: Not perceptible by human eyes; \textit{ii)} $1 < \Delta E \leq 2$: Perceptible only through close observation; \textit{iii)} $2 < \Delta E \leq 10$: Perceptible at a glance; \textit{iv)} $10 < \Delta E \leq 50$: Noticeably different colors; \textit{v)} $50 < \Delta E \leq 99$: Strong color difference; and \textit{vi)} $\Delta E = 100$: Completely different (opposite) colors. Figure~\ref{fig:teaser} illustrates eight results from our pipeline, showing samples from the CFD dataset that present the highest $\Delta E$ values across the four feature extraction methods, rendered under three lighting configurations for two ITA classes. As can be observed, both illumination and the extraction method influence the qualitative assessment.

\subsubsection{Individual Typology Angle (ITA) - Figure~\ref{fig:overview-model} (13).}

Although $\Delta E$ captures continuous perceptual error, it does not directly reflect the categorical classification of skin tone. To address this, we additionally calculated the ITA~\cite{chardon1991skin,merler2019diversity}, a dermatologically motivated metric that maps skin color to 
tone categories. 
The ITA is calculated from CIELAB values and allows for stratified analysis across six different skin tone groups. It is defined geometrically as an angle in the CIELAB color space, computed from the lightness ($L^*$) and yellow-blue chromaticity ($b^*$): $
\mathrm{ITA} = \arctan\left(\frac{L^* - 50}{b^*}\right) \times \frac{180}{\pi}.$ 
This formulation interprets skin color as a direction relative to a fixed reference point in the $(L^*, b^*)$ plane. Higher ITA angle values correspond to lighter skin tones, while lower or negative values correspond to darker skin tones. Using standard ITA thresholds, where Class I corresponds to ITA values greater than $55^\circ$, Class II to values between $41^\circ$ and $55^\circ$, Class III to the range from $28^\circ$ to $41^\circ$, Class IV to values between $10^\circ$ and $28^\circ$, Class V to values from $-30^\circ$ to $10^\circ$, and Class VI to ITA values lower than $-30^\circ$~\cite{chardon1991skin,merler2019diversity,feng2022fairness,feng2022towards}, we assigned both the original subject and the rendered MetaHuman to one of six skin tone categories. 
Inspired by the work of Feng et al.~\cite{feng2022fairness,feng2022towards}, for each rendered instance, we calculated the ITA Error as the absolute difference between the ITA value of the rendered MetaHuman and that of the corresponding reference image in CFD. Additionally, we assessed categorical precision by measuring whether the rendered character is assigned to the same ITA Class as the original subject, as discussed in the results. 

\section{Results}
\label{sec:results}

We present the results for both $\Delta E$ and ITA.
Statistical significance was assessed using Kruskal-Wallis tests with Dunn post hoc comparisons, as the normality and homoscedasticity 
were violated (in almost all cases: Shapiro-Wilk $p<.001$, Levene $p<.001$), and all tests considering significance level of 5\%. The results are presented separately for each ITA Class (I-VI) to allow for stratified equity analysis. It is important to notice that we did not find any significant 
results when comparing the texture generation methods in our pipeline (normalization and variation map). 

\begin{figure}
    \centering
    \includegraphics[width=0.9\linewidth]{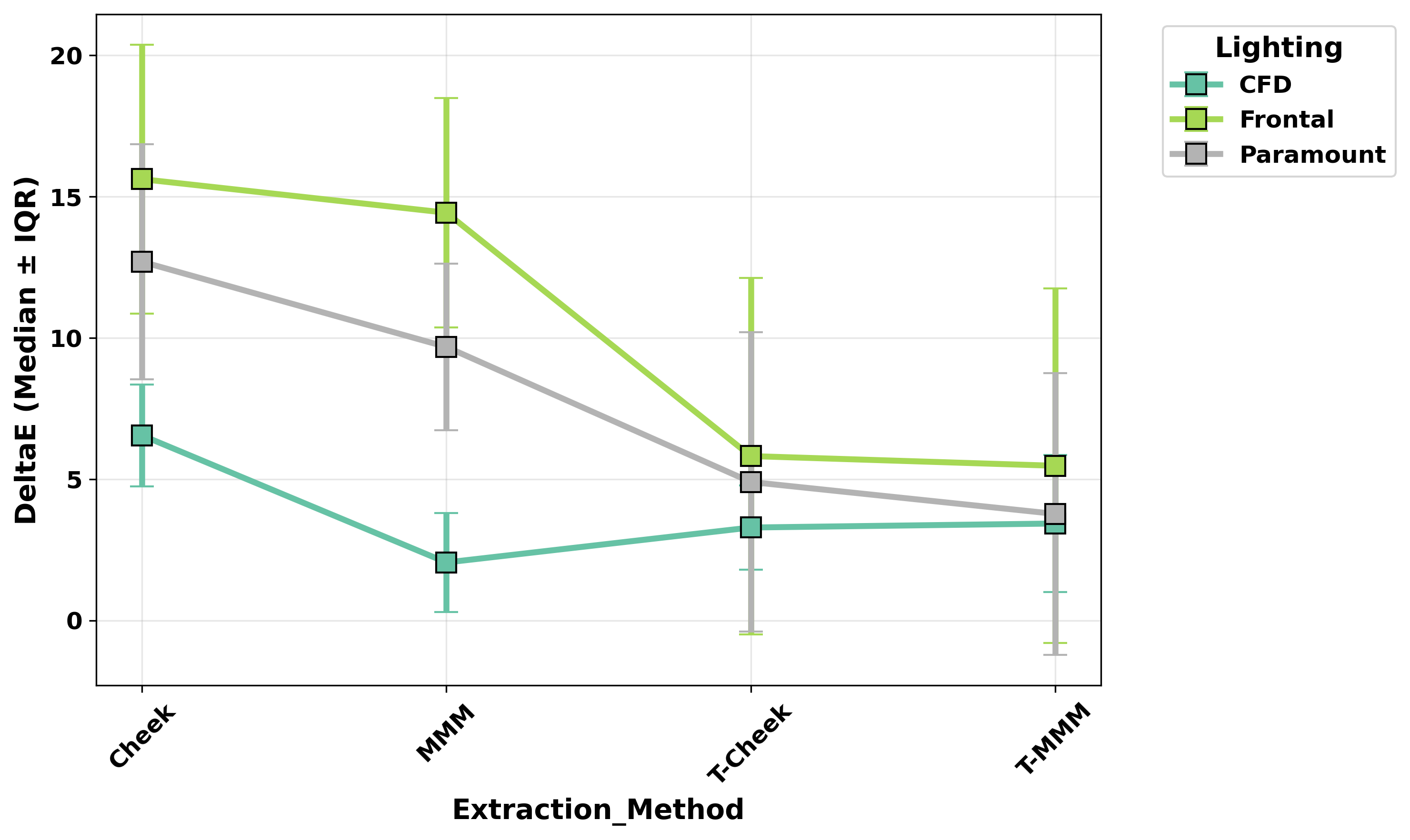}
    \caption{Values of $\Delta E$ for 3 light conditions and 4 extraction methods. Values of ITA Error considering 6 ITA Classes and 4 extraction methods. The higher the ITA Class number, the darker the skin tone.}
    \label{fig:DeltaE_MethodByLightingAndGeneral-extraction-method_1}
\end{figure}

\begin{figure}
    \centering
    \includegraphics[width=0.9\linewidth]{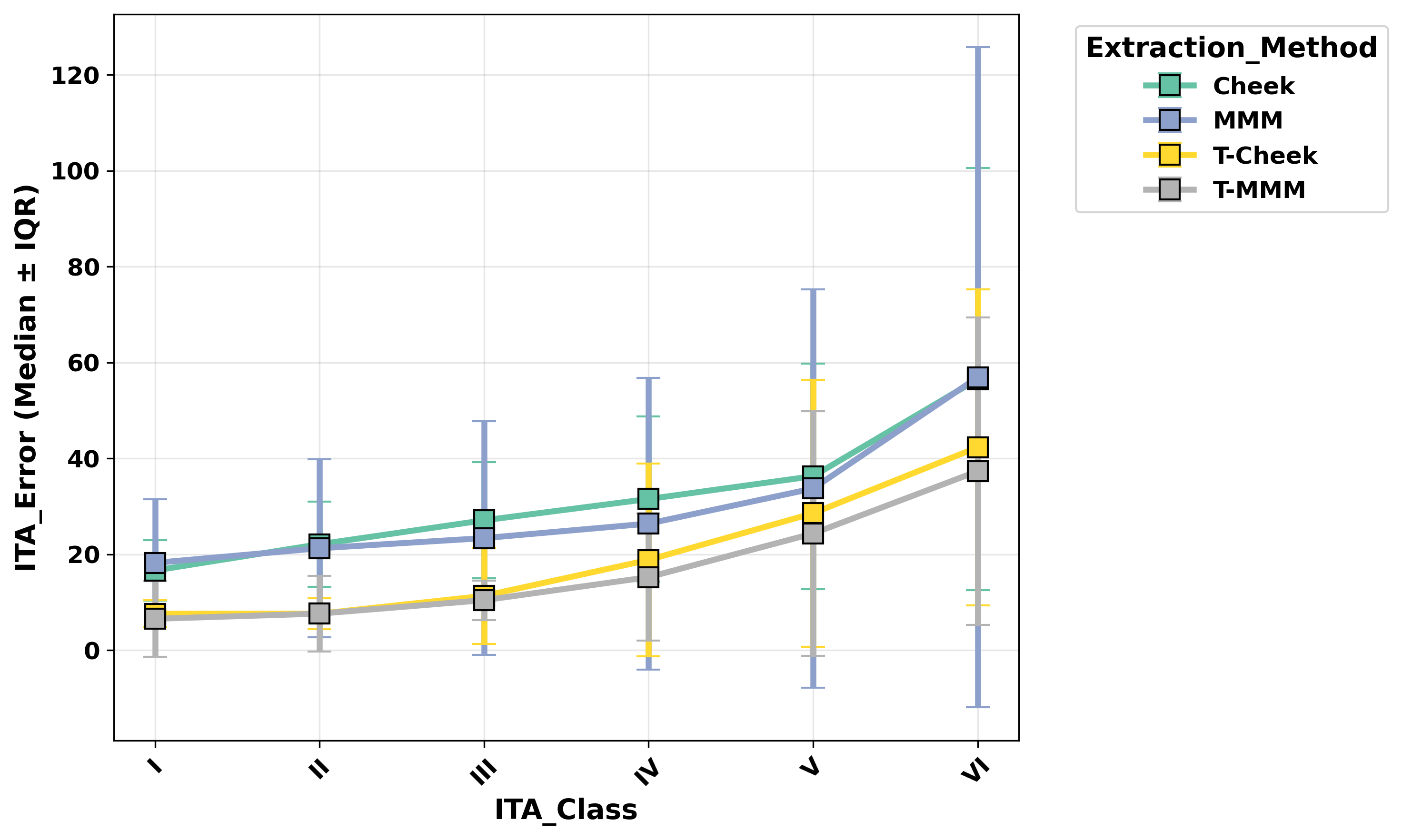}
    \caption{Values of ITA Error for 6 ITA Classes and 4 extraction methods. Values of ITA Error considering 6 ITA Classes and 4 extraction methods. The higher the ITA Class number, the darker the skin tone.}
    \label{fig:DeltaE_MethodByLightingAndGeneral-extraction-method_2}
\end{figure}

\begin{figure}
    \centering
    \includegraphics[width=0.9\linewidth]{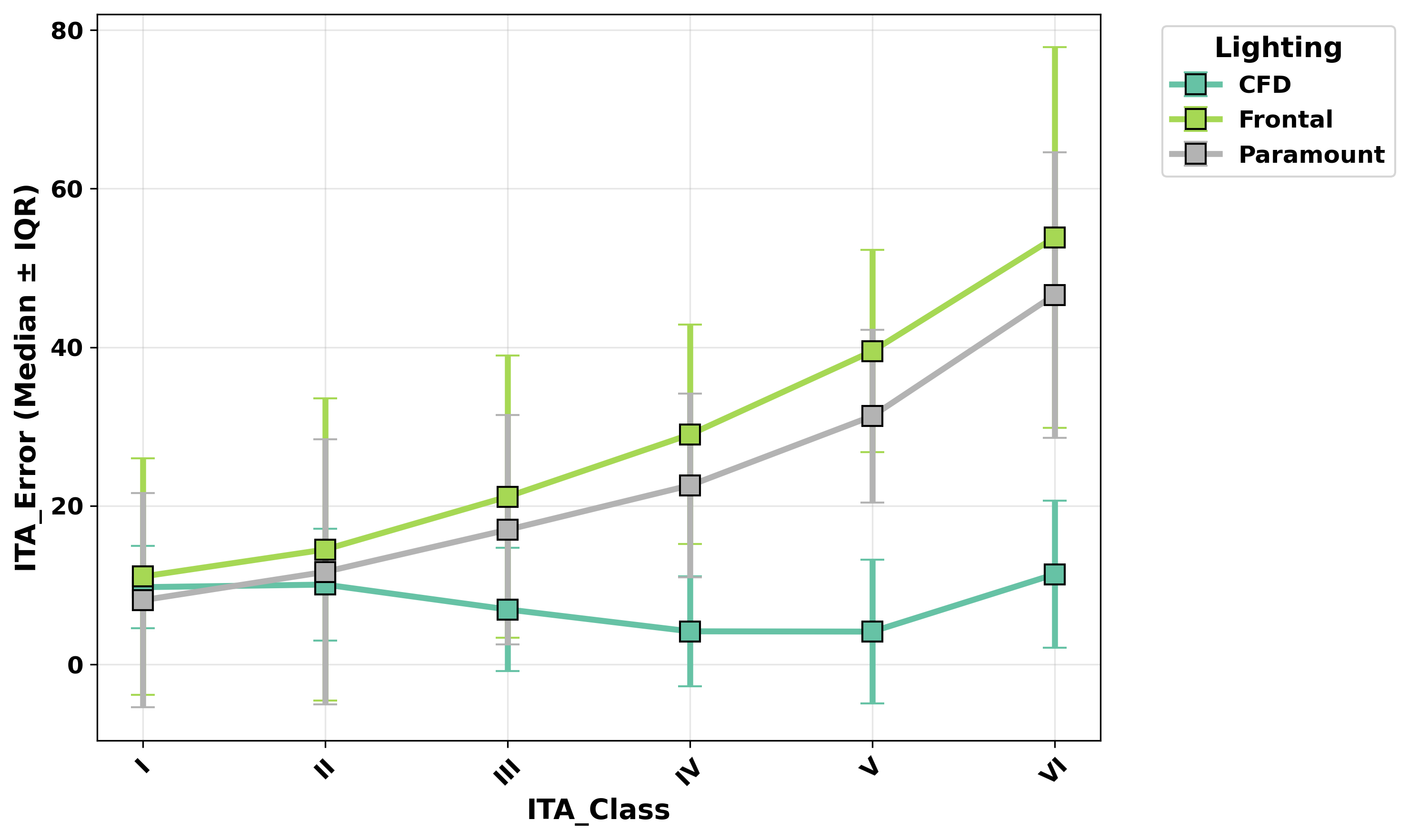}
    \caption{Values of ITA Error considering 6 ITA Classes and 3 light \newline conditions. The higher the ITA Class number, the darker the skin tone.}
    \label{fig:general_lightingAndGeneral_extraction_method_by_ligthing_1}
\end{figure}

\begin{figure}
    \centering
    \includegraphics[width=0.9\linewidth]{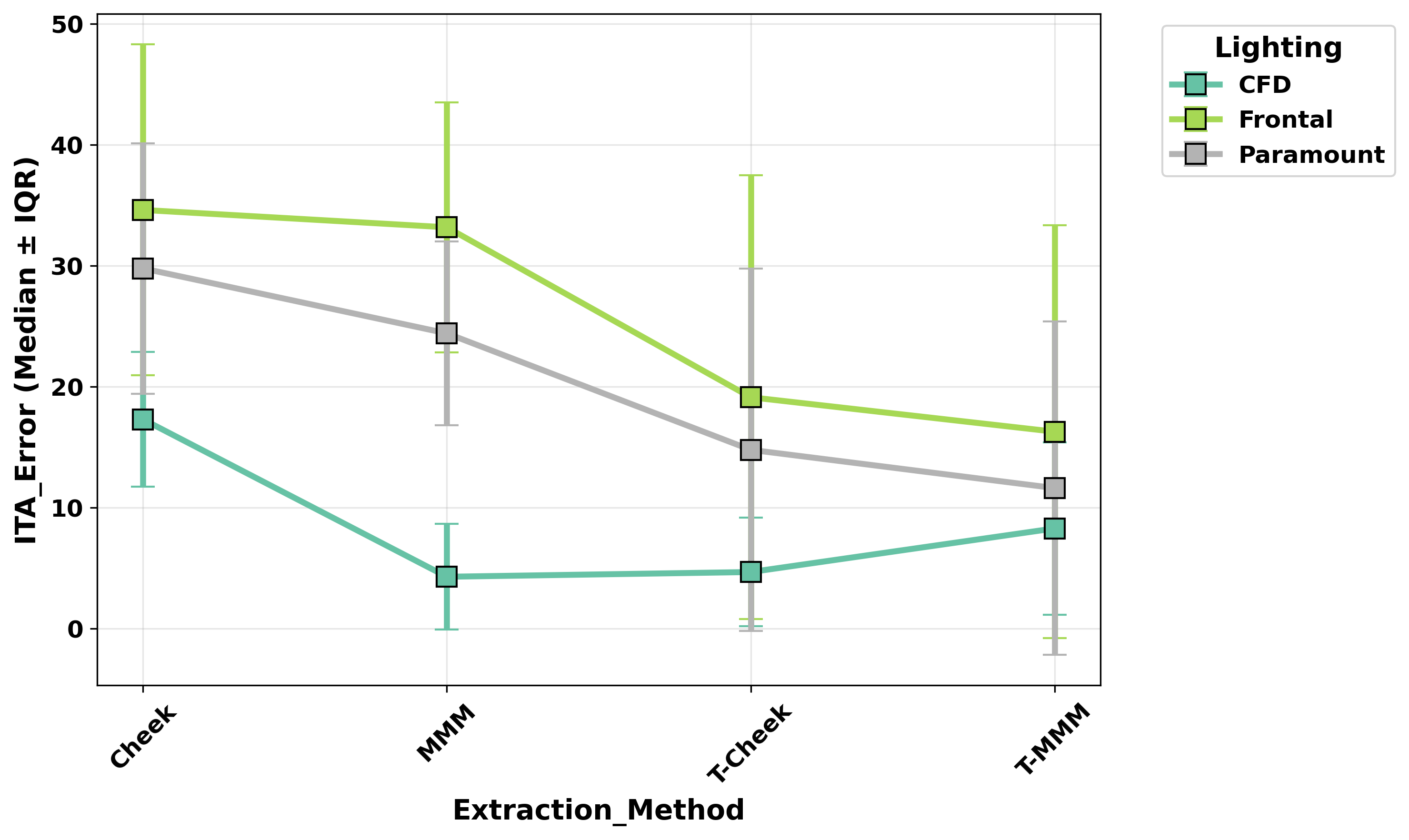}
    \caption{Values of ITA Error considering 3 light conditions and 4 extraction methods.}
    \label{fig:general_lightingAndGeneral_extraction_method_by_ligthing_2}
\end{figure}

\subsection{Perceptual Color Difference ($\Delta E$) Results}


Figure~\ref{fig:DeltaE_MethodByLightingAndGeneral-extraction-method_1} illustrates the $\Delta E$ values. As shown, the CFD Light, directly extracted from the images, yields lower overall error than the other light configurations, with the $T-Cheek$ and $T-MMM$ methods exhibiting the lowest average $\Delta E$ among the 4 extraction methods. The extraction method exhibited a main effect ($H = 5700.02$, $p < .001$). $Cheek$ (median=$12.69$) and $MMM$ (median=$9.67$) methods resulted in higher median errors than $T-$based variants ($T-Cheek$ with median=$4.90$, and $T-MMM$ with median=$3.77$), and all comparisons had significant differences ($p<.001$). 

Additionally, the results showed significant differences between the three evaluated light conditions (i.e., Frontal, Paramount, CFD)
($H=6504.39$, $p<.001$). CFD Light yielded the lowest errors (median $\Delta E \approx 2.2$-$4.5$), followed by Paramount Light (median $\Delta E \approx 5.8$-$14.6$), while Frontal Light consistently produced the highest perceptual error (median $\Delta E \approx 7.6$-$18.7$). All pairwise light comparisons were statistically significant ($p<.001$), confirming that illumination alone has a dominant impact on color fidelity. We also found an interaction effect between extraction methods and light conditions ($H=7219.64, p=<.001$). In addition, the comparisons between $T-Cheek$ and $T-MMM$ were the only ones that were statistically similar across all lighting conditions.

\begin{figure*}[!htb]
  \centering
  \includegraphics[width=14cm]{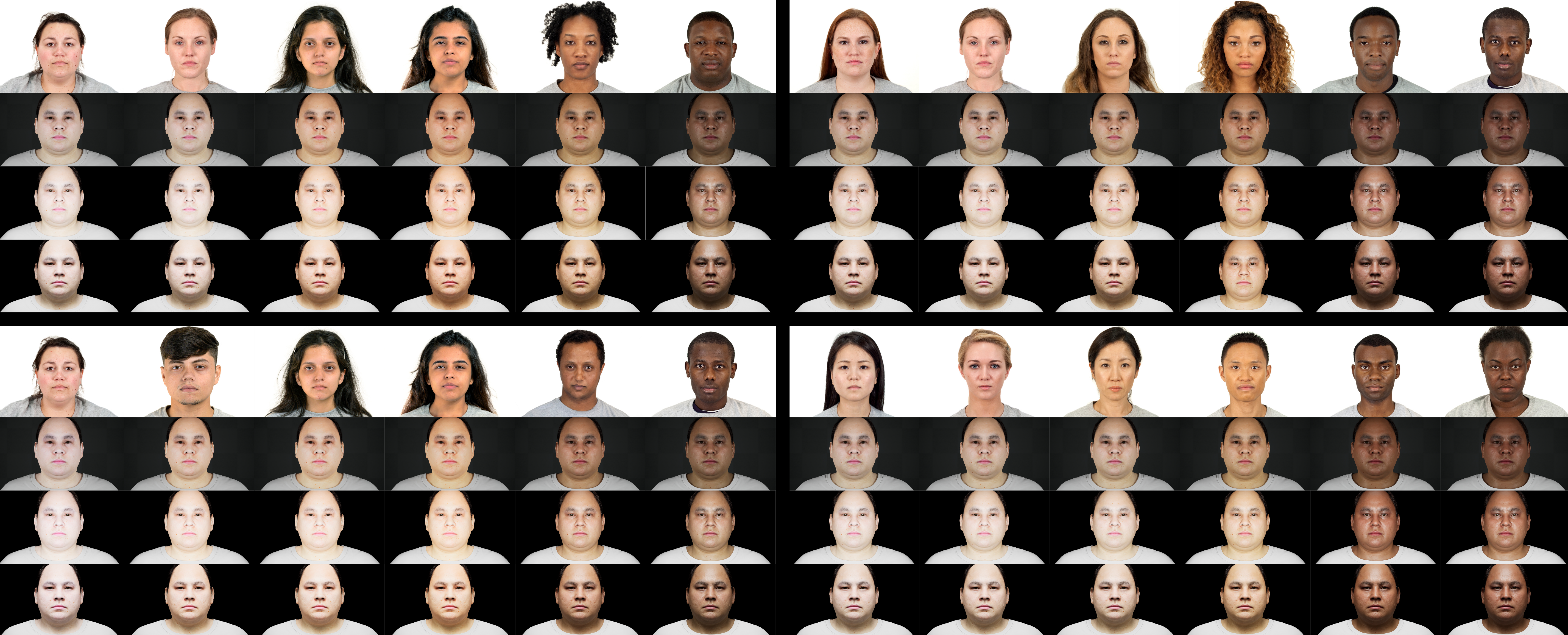}
  \caption{Quadrants show $MMM$, $T-MMM$, $Cheek$, and $T-Cheek$. Columns correspond to ITA I–VI Classes, and rows show the reference image (CFD) followed by renderings under CFD Light, Frontal Light, and Paramount Light. Samples represent the maximum color difference ($\Delta E$) for each ITA Class.}
  \label{fig:HighDeltaE}
\end{figure*}

\begin{figure*}[!htb]
  \centering
  \includegraphics[width=14cm]{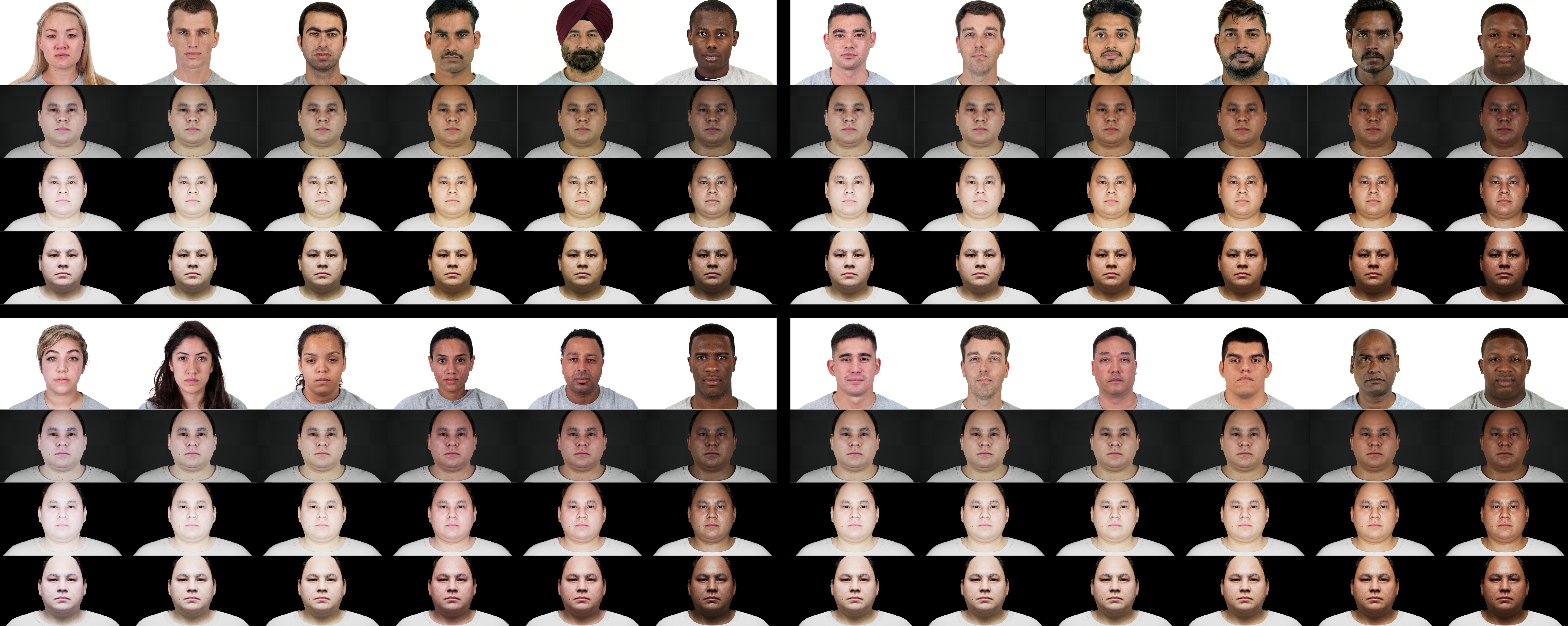}
  \caption{Quadrants show $MMM$, $T-MMM$, $Cheek$, and $T-Cheek$. Columns correspond to ITA I–VI Classes, and rows show the reference image (CFD) followed by renderings under CFD Light, Frontal Light, and Paramount Light. Samples represent the minimum color difference ($\Delta E$) for each ITA Class.}
  \label{fig:SmallDeltaE}
\end{figure*}

\begin{figure*}[!htb]
  \centering
  \includegraphics[width=8cm]{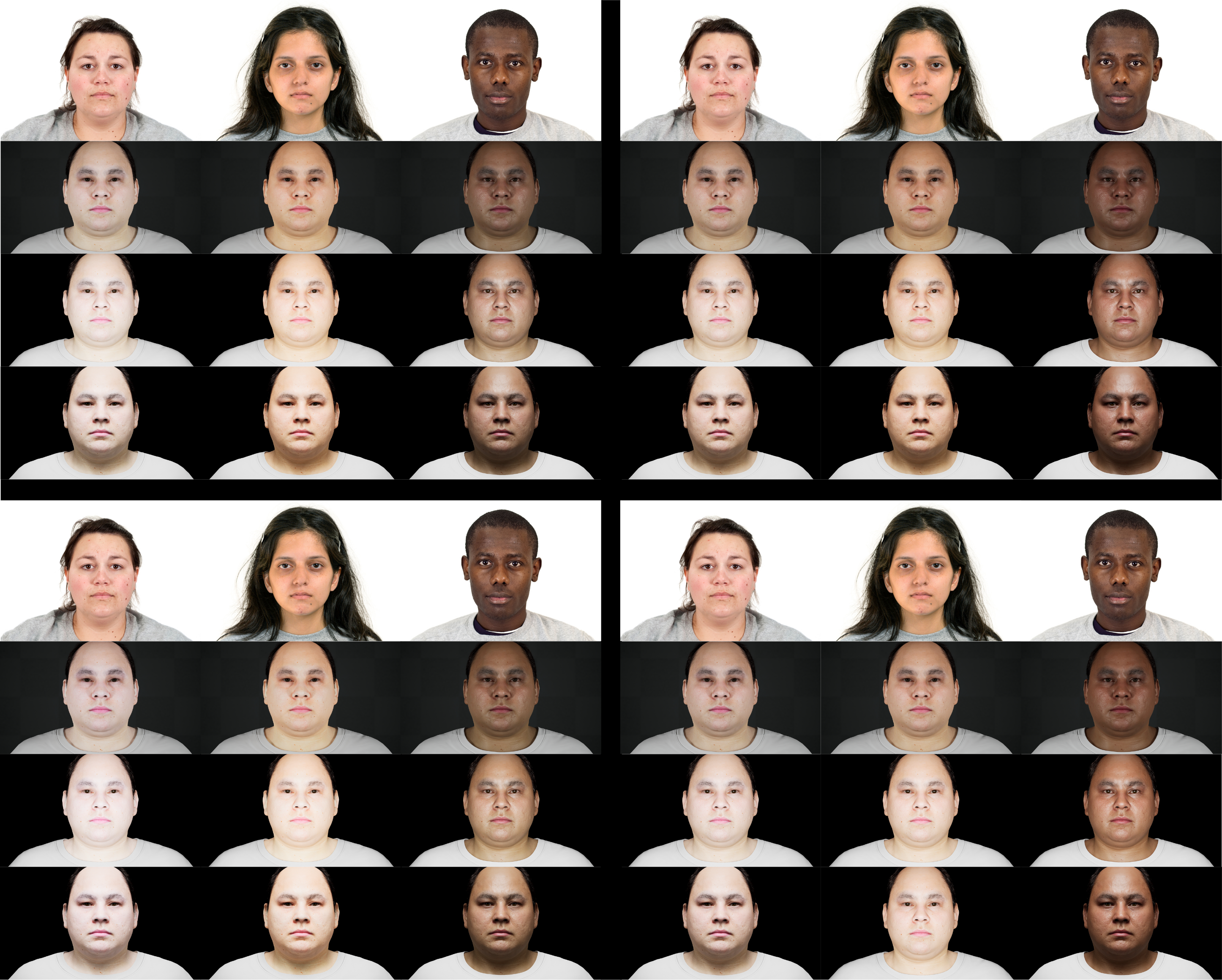}
  \caption{Quadrants show $MMM$, $T-MMM$, $Cheek$, and $T-Cheek$. Columns correspond to three subjects in the CFD dataset, and rows show the reference image (CFD) followed by renderings under CFD Light, Frontal Light, and Paramount Light. This figure represents ITA classes I, III, and VI, respectively.}
  \label{fig:VisualResults}
\end{figure*}

\subsection{Individual Typology Angle (ITA) Results}
\label{sec:ita_results}

The Chicago Face Database (CFD) provides self-reported demographic labels based on four ethnic categories (Asian, Black, Latino, and White). However, as our analysis focuses specifically on skin tone rather than ethnicity, we define a ground-truth labeling scheme based on skin color and do not use the CFD ethnic data.
For each reference image in CFD, we compute ITA values using the four color extraction strategies evaluated in our pipeline, namely $Cheek$, $MMM$, $T-Cheek$, and $T-MMM$. Each extraction method yields an independent ITA estimate and corresponding skin tone class. Given that these estimates may fall into different ITA Classes for the same subject, we define a single categorical ground truth skin tone label per image using a majority-vote criterion across the four extraction instances. In cases where no unique majority class is observed, ties are resolved by assigning the ITA Class obtained with the $T-MMM$ extraction method.
Of the 827 identities analyzed in the CFD, 81.38\% presented a dominant ITA Class per identity, defined as the one most frequently assigned among the four color extraction strategies applied to the same image. In the remaining 18.62\%, it was not possible to define a clear majority, as two or more ITA Classes occurred with equal frequency among the extractions. 
In these cases, the ITA Class was defined based on the $T-MMM$ strategy, which consistently selected the darkest class among those tied, with ITA III (29.87\%) and ITA V (26.62\%) being the most frequent resolutions. 
This choice is motivated by our hypothesis that $T-MMM$ provides the most robust skin tone estimation, as it reduces illumination effects while capturing color information from the full facial region. 

\subsubsection{Effect of Skin Color Extraction Method:}
\label{sec:results_extraction}

We statistically analyzed the difference between rendered and reference images as a function of skin tone category (six ITA Classes: I-VI) and extraction method (four types). Figure~\ref{fig:DeltaE_MethodByLightingAndGeneral-extraction-method_2} shows the ITA Error values for the 6 ITA Classes and 4 extraction methods. It can be noticed that the error values increase as the ITA Class represents darker skin tones (I - median=$12.12$; II - median=$14.45$; III - median=$17.35$; IV - median=$22.62$; V - median=$31.15$; VI - median=$49.43$). Extraction method exhibited a main effect ($H=3888.11$, $p<.001$). Non $T-$based methods produced higher errors than their counterparts ($p<.001$). Across all ITA Classes, the $Cheek$ and $MMM$ baselines yielded the largest median errors (respectively, medians of $29.32$ and $24.89$), while $T-Cheek$ and $T-MMM$ consistently reduced the error, with respectively medians of $15.08$ and $12.83$. All pairwise method comparisons were highly significant ($p<.001$), with the comparison between $T-Cheek$ and $T-MMM$ showing a small but also statistically significant difference ($p=.011$).

Beyond main effects, a significant interaction between ITA Class and extraction method was detected ($H=739.37$, $p<.001$), indicating that the impact of extraction method varies across skin tone classes. This interaction is visualized in Figure~\ref{fig:DeltaE_MethodByLightingAndGeneral-extraction-method_2}. For the $Cheek$ and $MMM$ methods, ITA Error increased with darker skin tones. For example, under $Cheek$ extraction, median error increased from $16.64$ (ITA I) to $56.54$ (ITA VI), while $MMM$ exhibited a similar result from $18.22$ to $56.93$. All within-method comparisons across ITA Classes were significant ($p<.001$), confirming a strong amplification of error for darker tones. In contrast, $T$-based methods attenuated, but did not eliminate, this trend. $T-Cheek$ reduced the median ITA VI error to $42.32$, and $T-MMM$ further reduced it to $37.34$, both significantly lower than their non-$T$ counterparts ($p<.001$). Importantly, while errors still increased with ITA Class under $T$-based extraction, the slope of this increase was visibly reduced, indicating a partial mitigation of skin-tone-dependent error accumulation. In addition, for lighter tones (ITA I-II), $T$-based methods reduced median error by approximately $9$-$14$ units relative to baseline methods, but without a significant effect. For darker tones (ITA V-VI), this reduction increased, reaching differences of over $15$-$20$ (e.g., ITA VI: $Cheek$ vs. $T-MMM$ median difference $>19$, $p<.001$). This pattern demonstrates that $T$-based extraction showed benefits for darker skin types in comparison with the other methods. Overall, these interaction effects indicate that differences are amplified for darker ITA Classes, reinforcing the importance of interaction-aware analysis when evaluating fairness-related performance across different skin tones.
Additionally, we evaluated the difference statistically among the ITA Classes to understand if the error is significantly higher in some classes. A main effect of ITA Class was observed ($H=2213.17$, $p<.001$). Median ITA Error increased with darker skin types, revealing a consistent effect. Median errors ranged from ITA I (median $\approx 7$-$18$ depending on method) to ITA VI (median $\approx 37$--$57$), with all pairwise comparisons between non-adjacent classes statistically significant (Dunn correction $p<.001$). Even adjacent classes showed significant differences (e.g., ITA I vs. II: $p<.001$), indicating the error's sensitivity to skin tone gradation.

\subsubsection{Impact of Lighting Conditions:}
\label{sec:results_lighting}

The results showed a significant main effect of lighting conditions ($H=7656.10$, $p<.001$). CFD Light resulted in the lowest categorical error (median=$8.34$), followed by Paramount Light (median=$19.78$), while Frontal Light produced the largest errors (median=$25.08$). All pairwise differences were statistically significant ($p<.001$). A significant interaction between ITA Class and lighting was detected ($H=5957.96$, $p<.001$), indicating an impact of illumination on ITA Class. As we can see in Figure~\ref{fig:general_lightingAndGeneral_extraction_method_by_ligthing_1}, for lighter skin tones (ITA I-II), 
ITA Error remained similar across lighting conditions. In contrast, for ITA III to VI, lighting had a non-linear effect. For example, ITA V exhibited a median of $4.14$ under CFD Light, which increased to $31.32$ under Paramount and rose to $39.51$ under Frontal Light. The most extreme effect was observed for ITA VI, where the median increased from $11.37$ (CFD) to $46.61$ (Paramount) and peaked at $53.85$ under Frontal Light. The differences between CFD and the other lights were significant (both $p<.001$). 
Furthermore, looking at Figure~\ref{fig:general_lightingAndGeneral_extraction_method_by_ligthing_2}, we can see that the CFD Light generated the fewest errors compared to the Frontal and Paramount Lights, especially for the $MMM$ and $T-Cheek$ methods, which showed statistically significant differences compared to the other two methods in Frontal and Paramount X CFD analysis (both $p<.001$).

\subsubsection{Skin Type Transition Analysis}

\begin{figure}[!htb]
  \centering
  \includegraphics[width=8cm]{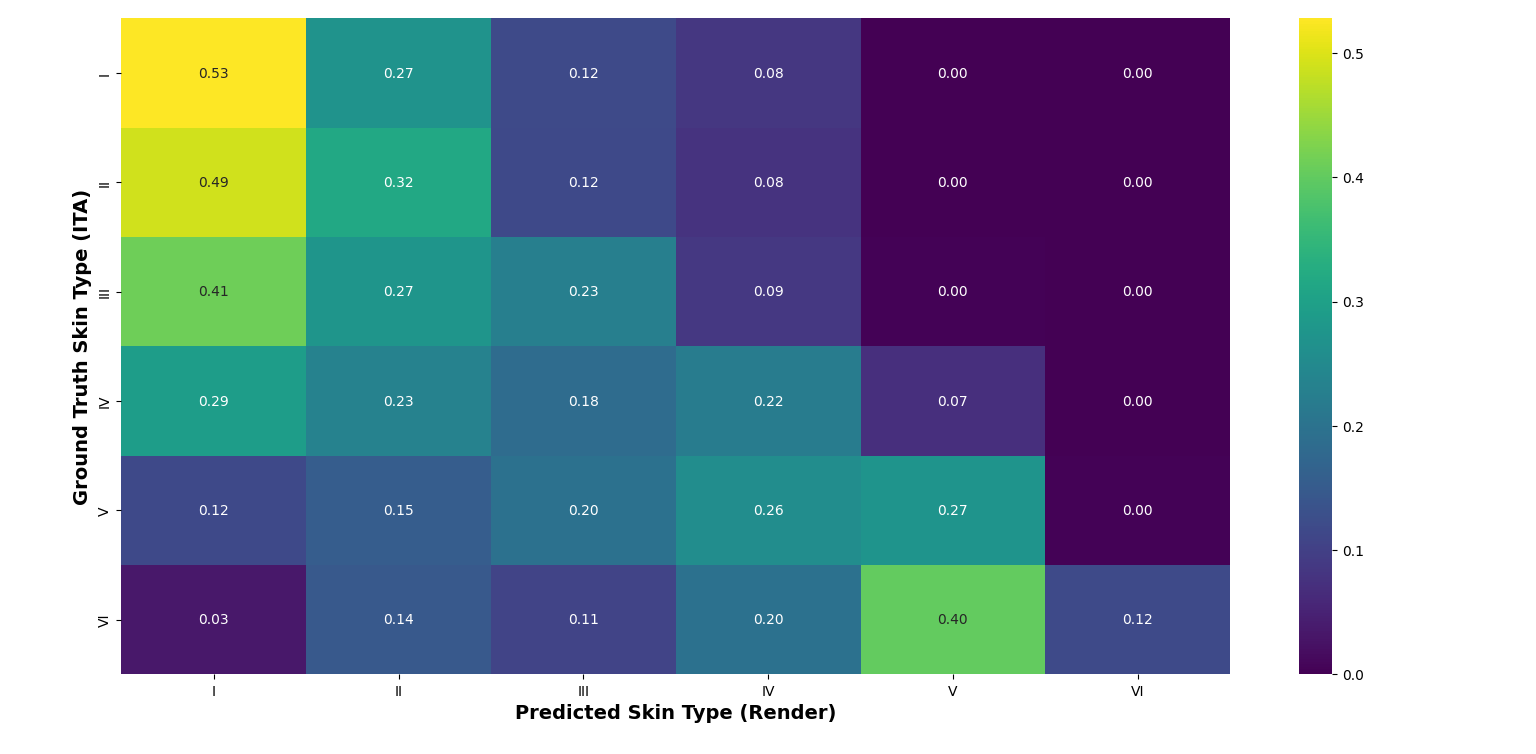}
  \caption{Confusion Matrix between ITA Class ground truth and ITA Class related to the Render.}
  \label{fig:general-matrix}
\end{figure}

To assess whether the rendering process alters the ITA Classes relative to the ITA ground truth, we constructed a skin class confusion matrix (as we can see in Figure~\ref{fig:general-matrix}) comparing the ITA Class, from the ground truth, with the ITA Class from the rendered image,
for all 19,848 images. 
In the results, the matrix showed that the lighter skin phototypes (I and II) exhibit partial preservation but already show dispersion towards neighboring classifications. The medium skin phototypes (III and IV) presented the most samples migrating to other classifications. For darker skin tones (V and VI), preservation rates are particularly low, with most samples being reassigned to lighter classifications. ITA VI is rarely preserved, indicating strong compression of darker phenotypes towards intermediate classes. 

\section{Discussion}


Firstly, it is important to note that, although this paper reports results using a single VH model, additional experiments were conducted with other models during the research phase, yielding no significant changes in the observed trends or conclusions.

Regarding question \textbf{Q1} (Extraction Methods Impact), our results show that the skin color extraction method plays a critical role in the fidelity of VH rendering, affecting both perceptual color difference ($\Delta E$) and ITA-based error. Methods relying solely on spatial regions or masking without illumination compensation, such as $Cheek$, produced errors exceeding the “perceptible difference” threshold reported in the literature~\cite{minaker2021optimizing,johnson2024biophysically}. In contrast, the $T$-based variants ($T$-Cheek and $T$-MMM) substantially reduced perceptual errors, remaining “perceptible at a glance”. While full-face multidimensional masking ($MMM$) provides a more holistic representation than the $Cheek$ region, illumination compensation ($T$) is the most critical factor for accuracy. As hypothesized before, the superior performance of $T$-MMM results from combining holistic full-face information with explicit illumination compensation, enabling a more faithful estimation of intrinsic skin reflectance.
Without this compensation, pipelines can propagate scene ambiguities (as identified by~\cite{feng2022fairness}) directly into the texture of a VH, leading to perceptible color distortions.

Regarding \textbf{Q2} (Phenotype Sensitivity), our results showed a significant effect of ITA Class, with error values systematically increasing as skin tone darkened from ITA I to ITA VI. This "error amplification" for darker phenotypes was greater with the $Cheek$ extraction method: the error for ITA VI was almost 4 times that for ITA I. The significant interaction between ITA Class and extraction method indicated that $T$-based methods can serve as a mitigation tool, even though $T-MMM$ did not eliminate the tendency for increased error among ITAs. 
These results corroborate the "over-lightening" bias described by Kim et al.~\cite{kim2022countering,kim2020racist},
which argued that the physical mechanism of subsurface scattering has been treated as synonymous with "human skin" rendering, despite being a dominant visual feature primarily for light skin phenotypes. Kim et al.~\cite{kim2022countering} identified that the fundamental subsurface scattering algorithms have been validated almost exclusively on white humans, leading to workflows that treat darker skin as a "deviation" from a white baseline. This disparity highlights the need for analyses that account for interactions to ensure fairness in VH generation.

The results for \textbf{Q3} (Lighting Influence) indicate that lighting is a dominant factor in color fidelity. Lighting conditions matching the source images (CFD) yielded the smallest errors, remaining within the “perceptible at a glance” range, whereas Frontal Light produced the largest perceptual errors. This aligns with Wisessing and McDonnell~\cite{wisessing2024blinded}, who show that although Frontal Light is common, it can intensify flatness and reduce contrast in virtual environments. Notably, the interaction analysis revealed that $T-Cheek$ and $T-MMM$ were the only methods to remain statistically consistent across lighting conditions, suggesting that illumination-compensated extraction provides a degree of lighting robustness. Overall, these findings reinforce that realism depends not only on the extracted color but also on its interaction with the virtual lighting environment, consistent with prior work~\cite{feng2022fairness,feng2022towards}.

Regarding the qualitative assessment, Figures~\ref{fig:HighDeltaE} and~\ref{fig:SmallDeltaE} present qualitative visual results illustrating the impact of skin color extraction methods and lighting conditions across the six ITA classes. The figures are organized into four quadrants: $MMM$ (top left), $T$-$MMM$ (top right), $Cheek$ (bottom left), and $T$-$Cheek$ (bottom right). Within each quadrant, columns progress from ITA~I to ITA~VI, from left to right. Rows, from top to bottom, show the reference image from the CFD dataset, followed by renderings under CFD, Frontal, and Paramount lighting conditions. The samples shown correspond to the cases with the largest color difference ($\Delta E$) for each ITA class in Figure~\ref{fig:HighDeltaE}, and to the smallest $\Delta E$ cases in Figure~\ref{fig:SmallDeltaE}. These examples support a qualitative assessment of how both illumination and extraction strategies influence the final appearance. Overall, the figures illustrate the combined effects of the three lighting configurations and the four extraction methods on skin tone reproduction. In particular, Figure~\ref{fig:VisualResults} highlights pronounced visual differences for three representative subjects from the CFD dataset, classified as ITA I, III, and IV, resulting from both the extraction method and the lighting setup. It can be observed that the $MMM$ and $Cheek$ methods (left quadrants) and the Frontal lighting condition (third row) tend to lighten the rendered skin tone, leading to visible distortions in skin tone reproduction. These qualitative observations are consistent and further corroborate the quantitative results.


\subsection{Limitations}
\label{sec:limitations}


Despite the strengths of the proposed methodology, some limitations should be acknowledged. First, our analysis relies on uncalibrated photographic inputs, which reflect common real-world avatar-creation scenarios but limit the recovery of physically accurate skin reflectance. Second, the evaluation is restricted to a single renderer (Unreal Engine) and a single VH asset (MetaHuman), which may influence the observed error patterns. Finally, our assessment is based solely on objective colorimetric metrics ($\Delta E$ and ITA-based error), which, while well established, do not fully capture subjective realism; user studies would be required to relate these errors to human perception.

\section{Final Considerations}
\label{sec:finals}

This work presented a fully automatic and scalable methodology for evaluating skin tone fidelity in VH generation pipelines. By integrating skin color and illumination extraction, texture re-colorization, real-time rendering, and objective colorimetric analysis, the proposed framework enables systematic and reproducible assessment of how skin appearance is affected by common design choices across the pipeline. Using large-scale experiments (19,848 images were generated and analyzed), we showed that both extraction strategies and rendering configurations significantly influence skin tone reproduction across different phenotypes. These results indicate that discrepancies in rendered skin appearance arise from interactions among multiple pipeline stages, rather than from isolated components. By explicitly modeling these interactions, our analysis reveals how biases and errors introduced at early stages of the pipeline can propagate to the final rendered output, providing a basis for benchmarking VH pipelines.
The objective of this work was not to recover physically accurate skin reflectance from photographic input, but to characterize the behavior of widely adopted extraction and rendering strategies when operating on non–color-calibrated images. From this perspective, the proposed methodology supports comparative evaluation and consistency analysis under realistic conditions. Future work should incorporate perceptual studies with human participants to establish a clearer link between computational image differences and subjective evaluations of realism, fairness, and visual comfort across different skin tones. Controlled experiments could help identify which rendering artifacts are perceptually salient and how they contribute to discomfort or bias. In parallel, machine learning approaches could be explored to automatically detect and mitigate skin-tone-related biases in rendering pipelines, using explainable models to reveal which visual features most influence perceptual disparities. 

\section{Acknowledgements}
\label{sec:Acknowledgements}

This study was partly financed by the Coordenação de Aperfeiçoamento de Pessoal de Nivel Superior – Brazil (CAPES) – Finance Code 001; by the Conselho Nacional de Desenvolvimento Científico e Tecnológico - Brazil (CNPq) - Process Numbers 309228/2021-2; 406463/2022-0; 153641/2024-0.

\bibliographystyle{sbc}
\bibliography{sbc-template}

\end{document}